# Adversarial Attacks on Image Classification Models: Analysis and Defense

Jaydip Sen
*Department of Data Science*
*Praxis Business School*
Kolkata, INDIA
email: jaydip.sen@acm.org

Abhiraj Sen
Department of CSE
*Jadavpur University*
Kolkata, INDIA
email: abhirajsen@ieee.org

Ananda Chatterjee
*Department of Data Science*
*Praxis Business School*
Kolkata, INDIA
email: ananda.chatterjee89@gmail.com

*Abstract*—The notion of adversarial attacks on image classification models based on convolutional neural networks (CNN) is introduced in this work. To classify images, deep learning models called CNNs are frequently used. However, when the networks are subject to adversarial attacks, extremely potent and previously trained CNN models that perform quite effectively on image datasets for image classification tasks may perform poorly. In this work, one well-known adversarial attack known as the fast gradient sign method (FGSM) is explored and its adverse effects on the performances of image classification models are examined. The FGSM attack is simulated on three pre-trained image classifier CNN architectures, ResNet-101, AlexNet, and RegNetY 400MF using randomly chosen images from the ImageNet dataset. The classification accuracies of the models are computed in the absence and presence of the attack to demonstrate the detrimental effect of the attack on the performances of the classifiers. Finally, a mechanism is proposed to defend against the FGSM attack based on a modified defensive distillation-based approach. Extensive results are presented for the validation of the proposed scheme.

## I. Introduction

In recent research, it has been observed that many machine learning models, including advanced neural networks, are vulnerable to what are called "adversarial samples" [1]. These are cases where the model makes incorrect predictions for inputs that are very similar to correctly classified ones. What's concerning is that various types of models built differently and trained on different data subsets, all tend to make the same mistakes in these adversarial examples. This suggests that there are underlying flaws in how we train these models. Initially, it was unclear why these adversarial cases occurred, but some theories suggested it might be due to the extreme non-linearity of deep neural networks, inadequate model averaging, or insufficient regularization during training.

However, a different perspective emerged from the work of Goodfellow et al. [2]. They argued that one does not necessarily need complex nonlinear models to create adversarial cases; linear behavior in high-dimensional spaces is sufficient. They also demonstrated that adversarial training could serve as a form of regularization for deep learning models, like techniques like dropout [3]. Switching to nonlinear model families like RBF networks was shown to reduce vulnerability to adversarial examples compared to standard regularization methods.

This vulnerability to adversarial attacks is particularly concerning in applications like autonomous vehicles, where deep neural networks are used to recognize road signs and other objects [4]. An attack that fools the system could lead to dangerous consequences [5-10].

Goodfellow et al. argue that there is a trade-off between building models that are easy to train (linear) and models that can withstand adversarial attacks (nonlinear) [2]. They suggest that developing better optimization methods could help strike a balance between these two aspects.

While most adversarial attacks focus on subtly altering image pixels, some attacks go beyond that. One approach involves creating an image-independent patch that, when placed on an image, tricks the classifier into making specific predictions [11]. These attacks can be more harmful because they do not require knowledge of the specific image being targeted.

This paper delves into various adversarial attacks on image classification models, with a particular focus on the *fast gradient sign method* (FGSM) and the adversarial patch attack. It examines how these attacks impact image classification accuracy and presents detailed results.

The main contribution of the current work is threefold. First, it introduces the concepts of adversarial attacks on image classification models and presents two different types of attacks, the *fast gradient sign method* (FGSM) and patch attack. Second, it demonstrates how these attacks can adversely affect the classification performance of deep neural networks which are otherwise (i.e., in the absence of the attacks) very accurate in classifying images from large databases of images. Finally, it introduces the concept of defensive distillation and shows how the FGSM attacks can be defended using the distillation approach.

The rest of the paper is organized as follows. Section II presents some related works on adversarial attacks. Some theoretical background information on adversarial attacks and pre-trained image classification models is discussed in Section III. Section IV presents detailed results of the attacks and their analysis. Section V presents a discussion on distillation and how it has been used for defending against adversarial attacks. Section VI presents the results of the distilled network as a defense against attacks. Finally, the paper is concluded in Section VII highlighting some future works.

## II. Related Work

Deep learning systems are generally susceptible to what are known as adversarial instances. These are carefully chosen inputs that cause the network to modify its output in a way that is not easily noticeable to a human observer [5, 17]. Various optimization techniques, such as L-BFGS [1], the Fast Gradient Sign Method (FGSM) [2], DeepFool [12], and Projected Gradient Descent (PGD) [13], can be employed to discover these adversarial examples. Typically, these changes affect individual pixels by only a small amount. Some attack approaches focus on altering just a small portion of an image's pixels, as seen in methods like Jacobian-based saliency maps [14], or they target a small patch at a predefined location [15].

Szegedy et al. conducted a study that unveiled numerous intriguing characteristics of neural networks and related models [1]. Here are some notable findings from their research. (i) The study consistently demonstrated that the box-constrained L-BFGS method reliably identifies adversarial cases. (ii) Adversarial instances within the ImageNet dataset [16] closely resemble the original examples to such an extent that distinguishing between the two is practically impossible for a human observer. (iii) A common trend observed was that the same adversarial example was often incorrectly classified by a significant number of classification models, each of which had been trained using a distinct subset of the training data. (iv) In some instances, adversarial events were found to decrease the robustness of shallow Softmax regression models. (v) It was noted that training models on adversarial examples could effectively enhance the regularization of classification models.

Eykholt et al. illustrated a range of methods for misleading machine learning models by either producing large posters resembling stop signs or affixing different stickers to actual stop signs [17].

These findings indicate that classifiers developed through contemporary machine learning techniques often fail to grasp the fundamental principles governing correct label assignments, even if they exhibit outstanding performance on test data. While these models excel when dealing with naturally occurring data, their classification accuracy significantly diminishes when confronted with data points possessing a low probability within the underlying data distribution. This presents a significant challenge in the realm of image classification, as convolutional neural networks employed for classification rely on computing perceptual feature similarity based on Euclidean distance. However, this method can yield false resemblances when images with unrealistically small perceptual differences are assigned to different classes according to the neural network's representation.

The issue described above is especially pertinent to deep neural networks, although linear classifiers can also be affected by it. No model has managed to remain impervious to adversarial perturbations while maintaining top-tier accuracy on clean inputs so far. Nevertheless, researchers have put forth various methods for defending against adversarial attacks based on small perturbations, along with some innovative training techniques [13, 14, 18, 19-27].

Some of these works proposing methods to defend against adversarial attacks are briefly presented in the following.

Madry et al. developed and trained deep neural networks using the MNIST and CIFAR10 image datasets, ensuring their resilience against various adversarial attacks [13]. They devised a method to pinpoint a saddle point for error function optimization and employed a projected gradient descent (PGD) as the adversary. This approach achieved a noteworthy classification accuracy of 89% when tested against the most formidable adversary.

Papernot et al. introduced an innovative approach for generating adversarial samples by gaining a deep understanding of how inputs and outputs are linked in deep neural networks [14]. In a computer vision scenario, the researchers demonstrated that their method could consistently produce samples that were accurately classified by humans but misclassified by a deep neural network, achieving a 97% success rate in adversarial attacks while only altering an average of 4.02% of the input characteristics for each sample. Additionally, they devised a measure of difficulty to evaluate the vulnerability of different sample classes to adversarial perturbations and proposed a defense strategy against such adversarial samples.

Tramer et al. noted that adversarial attacks have a more significant impact in a black-box scenario, where perturbations are calculated and applied to unprotected models [18]. These attacks are highly effective when executed in a single step, allowing them to quickly escape the non-smooth vicinity of input data through a brief random adjustment. The authors introduced a technique called ensemble adversarial training, which involves incorporating perturbations obtained from other models into the training data. This approach has demonstrated resistance to black-box adversarial attacks on the ImageNet dataset.

Dong et al. established a dependable benchmark for evaluating the resilience of image classification models against adversarial attacks [21]. They made several noteworthy observations, including the following: Firstly, adversarial training stands out as one of the most effective defense strategies due to its ability to adapt to various threat models. Secondly, the use of model robustness curves proves valuable in assessing how well models withstand adversarial challenges. Lastly, defenses that rely on randomization demonstrate greater resistance when confronted with query-based black-box attacks.

Chen et al. scrutinized and assessed multiple defense strategies for their features and effectiveness against adversarial attacks [22]. The authors conducted their evaluation from four distinct angles, encompassing (i) gradient masking, (ii) adversarial training, (iii) detection of adversarial examples, and (iv) modifications to input data. They highlighted the advantages and limitations of various defense mechanisms in the context of adversarial attacks and also delved into prospective directions for designing more robust methods to protect image classification models against such threats.

III. BACKGROUND THEORIES

In this section, theoretical background information is provided for the understanding of the readers. The concepts such as adversarial attacks and the fast gradient sign method (FGSM) attack, as well as three pre-trained deep neural network models based on convolutional neural networks (CNNs): ResNet-104, AlexNet, and RegNetY 400MF.

**(1) *Adversarial attacks:*** Numerous adversarial attack strategies have been devised, all with the common objective of significantly altering a model's prediction by making slight modifications to the input data or image. How can we manipulate the image of a goldfish in a way that a classification model, which could initially correctly identify it as a goldfish, would no longer do so? Meanwhile, a human observer would still unequivocally recognize the image as a goldfish, so the label of the image should remain unchanged. The primary objective of a generator network within the framework of generative adversarial networks (GANs) aligns with this concept: it aims to deceive another network (a discriminator) by manipulating its input.

**(2) *Fast Gradient Sign Method:*** The Fast Gradient Sign Method (FGSM), initially introduced by Ian Goodfellow and his colleagues, stands as one of the earliest attack techniques [2]. FGSM employs a neural network's gradients to craft an adversarial image. Essentially, this approach generates an adversarial image by calculating the gradients of a loss function (such as mean-square error or categorical cross-entropy) concerning the input image and then utilizing the sign of these gradients to create a new image, known as the adversarial image, which maximizes the loss. The outcome is an output image that, to the human eye, appears identical to the original, but it causes the neural network to anticipate something different from what it should have. The FGSM can be represented as in (1).

$$adv_x = x + \varepsilon * sign(\nabla_x J(\theta, x, y)) \qquad (1)$$

In (1) $adv_x$ is the adversarial image as the output, $x$ is the original image as the input, $y$ is the actual class (i.e., the ground-truth label) of the input image, $\varepsilon$ is the noise intensity expressed as a small fractional value by which the signed gradients are multiplied to create perturbations, $\theta$ is the neural network model used for image classification, and $J$ is the loss function.

The FGSM attack applied to an image consists of three sequential steps. First, the loss function's value is calculated following the forward propagation within the network. Next, gradients are computed concerning the pixels within the original (i.e., input) image. Finally, the input image's pixels are subtly adjusted in the direction of the computed gradients to maximize the value of the loss function.

In machine learning, starting by calculating the loss after the forward propagation step is common practice. A negative likelihood loss function is typically employed to assess how closely the model's prediction aligns with the actual class. Gradients come into play when determining the direction to adjust the weights to minimize the loss function during neural network training. However, calculating gradients for the pixels of an image is not a standard task. In the case of FGSM, the pixels within the input image are shifted in the gradient's direction to maximize the loss function's value.

 **(3) *ResNet101 architecture:*** ResNet-101 is a variant of the ResNet (Residual Neural Network) architecture, known for its depth and use of residual blocks with skip connections [28]. With 101 layers, ResNet-101 is a deep neural network commonly used for image-related tasks such as image classification, object detection, and image segmentation. Its architecture, featuring skip connections, helps mitigate training challenges associated with deep networks. ResNet-101 is known for its use of skip connections or residual connections. The skip connections allow information to flow more easily through the network by adding shortcut connections that bypass one or more layers. Pre-trained versions of ResNet-101 on large datasets are frequently used for transfer learning. Transfer learning involves taking a pre-trained model (trained on a large dataset like ImageNet) and fine-tuning it on a specific task with a smaller dataset. This approach can save significant training time and computational resources. Overall, ResNet-101 is recognized for achieving state-of-the-art performance in various computer vision applications including image classification, object detection, and image segmentation. Its depth and skip connections make it suitable for capturing intricate features in images.

(4) *Alexnet architecture:* AlexNet is a deep convolutional neural network (CNN) architecture that achieved remarkable success in computer vision tasks, particularly image classification [29]. It consists of eight layers, including five convolutional layers, followed by three fully connected layers. AlexNet's key features include the use of ReLU activation functions, max-pooling layers, dropout regularization, and local response normalization (LRN). It won the ImageNet Large Scale Visual Recognition Challenge (ILSVRC) in 2012 and played a significant role in popularizing deep learning for computer vision applications, serving as a foundational model for subsequent CNN architectures.

(5) *RegNetY 400MF architecture:* RegNetY 400MF is a member of the RegNet family of neural network architectures [30]. It is designed to provide efficient and scalable solutions for various computer vision tasks. The "400MF" in its name signifies that it has around 400 million parameters, making it a balanced choice between model size and performance. RegNet architectures aim to achieve competitive performance on computer vision tasks while being computationally efficient. They achieve this through carefully designed network blocks and regularization techniques. Due to their scalability and efficiency, RegNet models can be used for transfer learning, where a pre-trained model on a large dataset (e.g., ImageNet) can be fine-tuned for specific tasks with smaller datasets. RegNetY 400MF is used for tasks such as image classification, object detection, and semantic segmentation and can be fine-tuned for specific applications. It stands out for its efficiency, scalability, and versatility in the field of computer vision.

IV. RESULTS OF FGSM AND PATCH ATTACKS

The experiments were carried out to investigate the impact of two distinct types of adversarial attacks on three widely recognized pre-trained CNN architectures. The study involved applying the FGSM attack and the adversarial patch attack to a set of images. These attacks were simulated on three pre-trained architectures: ResNet-101, AlexNet, and RegNetY 400MF. The images were selected from the ImageNet dataset [16], and the pre-trained CNN models (ResNet-101, AlexNet, and RegNetY 400MF) available in PyTorch's torchvision package were utilized in the experiments.

**(1) *Classification in the absence of an attack:*** Before delving into the examination of how adversarial attacks affect image classification models, we first assess the classification accuracy of these models in the absence of any attack. Given that the ImageNet dataset comprises 1000 classes, it is not prudent to solely evaluate a model's performance based on its classification accuracy. For instance, consider a model that consistently predicts the true label of an input image as the second-highest class using the Softmax activation function. While it does recognize the object in the image, its accuracy would be recorded as zero. The reason behind this is that there is not always a single, distinct label that can be assigned to an image among ImageNet's 1000 classes. This is why "Top-5 accuracy" is a widely used alternative metric for image classification across many classes. It indicates how frequently the true label falls within the model's top 5 most likely predictions. Given the strong performance of the three pre-trained architectures on ImageNet images, the results present error rates (1 - accuracy) rather than accuracy values.

TABLE I. CLASSIFICATION ACCURACY OF THE CNN MODELS ON THE IMAGENET DATA

| Metric | ResNet101 Model | AlexNet Model | RegNetY 400MF Model |
|---|---|---|---|
| Top-1 error | 13.30% | 28.06% | 18.72% |
| Top-5 error | 2.16% | 9.90% | 3.76% |

Table I displays the performance outcomes of three classification models across the complete ImageNet dataset, which encompasses 1000 different image classes. All three models exhibit a high level of accuracy, as indicated by their low Top-% error percentages. Among these architectures, the ResNet-101 model stands out with the highest accuracy and the lowest error rates. Specifically, the Top-5 and Top-1 error rates for this model are 2.16% and 13.30%, respectively.

TABLE II. THE CLASSIFICATION RESULTS OF THE RESNET-101 MODEL FOR THE CHOSEN IMAGES

| Image Index | Image True Class | Top-5 Predicted Classes and Confidences | |
|---|---|---|---|
| | | Class | Confidence |
| 7 | goldfish | **goldfish** | **0.9996** |
| | | tench | 0.0002 |
| | | flamingo | 0.0001 |
| | | black swan | 0.0001 |
| | | terrapin | 0.0001 |
| 16 | tiger shark | **tiger shark** | **0.5778** |
| | | electric ray | 0.1993 |
| | | great white shark | 0.1011 |
| | | hammerhead | 0.0783 |
| | | stingray | 0.0176 |
| 20 | hammerhead | **hammerhead** | **0.7796** |
| | | tiger shark | 0.1776 |
| | | electric ray | 0.0297 |
| | | great white shark | 0.0042 |
| | | gar | 0.0029 |
| 27 | electric ray | **electric ray** | **0.9985** |
| | | stingray | 0.0008 |
| | | loggerhead | 0.0005 |
| | | leatherback turtle | 0.0001 |
| | | dugong | 0.0001 |

Following the overall evaluation of the three models on the ImageNet dataset, we proceed to examine specific images within the dataset. To do this, we randomly select images with the indices 7, 16, 20, and 27 and assess how the models have classified these images. Notably, the images corresponding to these four indices belong to the classes "goldfish," "tiger shark," "hammerhead" and "electric ray", respectively.

Table II displays the ResNet-101 model's performance in classifying the four specified images. The results show that the model's classification accuracy is exceptionally high, with confidence levels exceeding 90% for the true class except for the image "tiger shark". In this context, "confidence" refers to the probability value assigned by the model to the corresponding class. For instance, when the ResNet-101 model correctly predicts the class "goldfish" for an image, it associates a confidence value of 0.9996 with this classification, indicating a high probability (99.96%) that the image belongs to the "goldfish" class.

Fig. 1 illustrates the ResNet-101 model's classification results for the four images. Within Figure 1, the input image is displayed on the left, while on the right, the model's confidence values are listed for the top five predicted classes for that image. These confidence values are represented as horizontal bars.

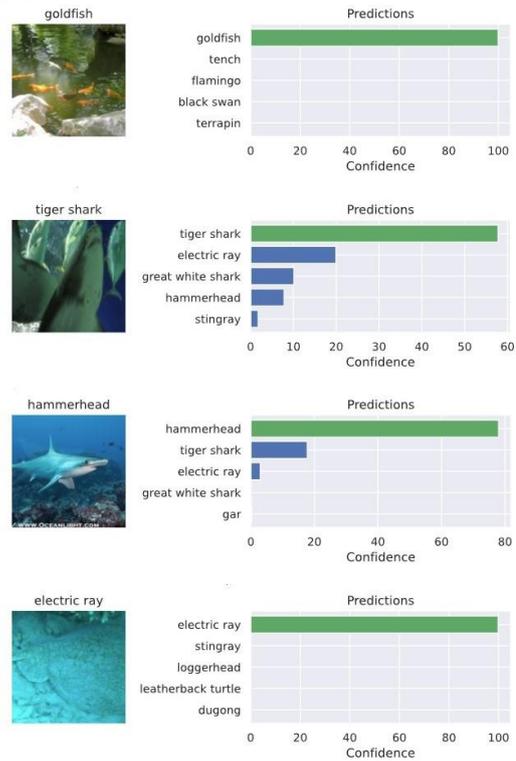

Fig. 1. The classification of some specific images by the ResNet-101 model.

**(2)** *Classification in the presence of the FGSM attack:* After assessing the performance of the three CNN architectures in the context of image classification tasks using the ImageNet dataset, we delve into examining the impact of adversarial attacks on these classifier models. Our initial focus is on the FGSM attack with an epsilon (ε) value of 0.02. This ε value of 0.02 implies that the pixel values are altered by an amount of approximately 1 within the range of 0 to 255, which covers the full range of potential pixel value changes. This alteration is so subtle that distinguishing the adversarial image from the original is practically impossible. We present the performance outcomes of the ResNet-101 when subjected to the FGSM attack with ε = 0.02 in Table III. Additionally, the results are visually represented in Fig. 2.

TABLE III. THE CLASSIFICATION RESULTS OF THE RESNET-101 MODEL UNDER FGSM ATTACK WITH EPSILON = 0.02 FOR CHOSEN IMAGES

| Image True Class | Top-5 Predicted Classes and Confidence | |
|---|---|---|
| | Class | Confidence |
| **goldfish** | hippopotamus | 0.2372 |
| | lakeside | 0.1076 |
| | terrapin | 0.0555 |
| | tench | 0.0497 |
| | beaver | 0.0485 |
| **tiger shark** | electric ray | 0.8774 |
| | stingray | 0.0782 |
| | hammerhead | 0.0053 |
| | killer whale | 0.0046 |
| | dugong | 0.0039 |
| **hammerhead** | tiger shark | 0.7204 |
| | electric ray | 0.2166 |
| | **hammerhead** | 0.0244 |
| | coral reef | 0.0147 |
| | stingray | 0.0043 |
| **electric ray** | loggerhead | 0.7381 |
| | leatherback | 0.1179 |
| | **electric ray** | 0.1038 |
| | conch | 0.0143 |
| | stingray | 0.0090 |

The impact of the FGSM attack on the ResNet-101 classifier model is quite adverse, even when using a low ε value of 0.02, while it is challenging to distinguish the adversarial images from the original ones.

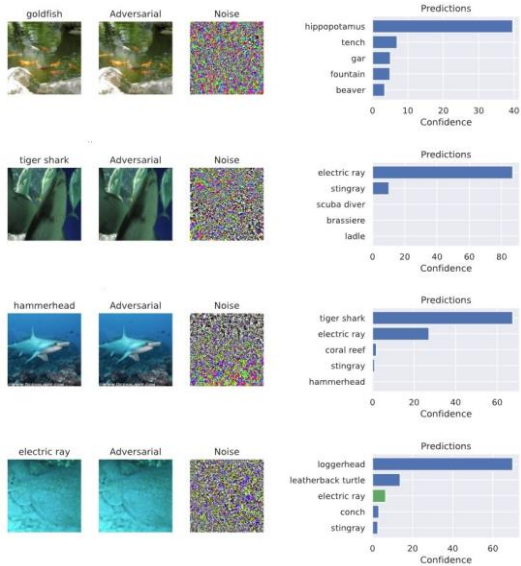

Fig. 2. The classification results of the ResNet-101 model under the FGSM attack with ε = 0.02.

TABLE IV. THE CLASSIFICATION PERFORMANCE OF RESNET-101 MODEL UNDER FGSM ATTACK FOR DIFFERENT VALUES OF EPSILON

| Noise Level (ε) | Top-1 Error (%) | Top-5 Error (%) |
|---|---|---|
| 0.01 | 77.18 | 33.34 |
| 0.02 | 87.10 | 48.78 |
| 0.03 | 90.50 | 55.62 |
| 0.04 | 91.58 | 58.98 |
| 0.05 | 92.02 | 60.82 |
| 0.06 | 92.20 | 61.60 |
| 0.07 | 92.02 | 61.96 |
| 0.08 | 91.88 | 62.16 |
| 0.09 | 91.66 | 61.84 |
| 0.10 | 91.06 | 61.80 |

TABLE V. THE CLASSIFICATION PERFORMANCE OF ALEXNET MODEL UNDER FGSM ATTACK FOR DIFFERENT VALUES OF EPSILON

| Noise Level (ε) | Top-1 Error (%) | Top-5 Error (%) |
|---|---|---|
| 0.01 | 79.04 | 44.16 |
| 0.02 | 92.28 | 65.30 |
| 0.03 | 96.36 | 75.92 |
| 0.04 | 98.02 | 81.78 |
| 0.05 | 98.80 | 85.84 |
| 0.06 | 99.10 | 88.34 |
| 0.07 | 99.24 | 89.98 |
| 0.08 | 99.34 | 91.02 |
| 0.09 | 99.38 | 92.18 |
| 0.10 | 99.48 | 92.84 |

TABLE VI. THE CLASSIFICATION PERFORMANCE OF REGNETY 400MF MODEL UNDER FGSM ATTACK FOR DIFFERENT VALUES OF EPSILON

| Noise Level (ε) | Top-1 Error (%) | Top-5 Error (%) |
|---|---|---|
| 0.01 | 89.56 | 53.12 |
| 0.02 | 95.12 | 66.16 |
| 0.03 | 96.42 | 71.44 |
| 0.04 | 96.80 | 73.98 |
| 0.05 | 97.00 | 75.30 |
| 0.06 | 97.08 | 75.68 |
| 0.07 | 96.74 | 75.88 |
| 0.08 | 96.68 | 76.04 |
| 0.09 | 96.50 | 75.86 |
| 0.10 | 96.24 | 75.72 |

The parameter ε is incrementally increased from 0.01 to 0.10 in steps of 0.01. It is observed that, with a few exceptions, the classification error consistently rises with ε until it reaches a value within the range of 0.07 to 0.09. The impact of the FGSM attack is so significant that the classification error for the ResNet-101 model in the presence of this attack reaches notably high levels, reaching 92.20% for Top-1 error and 62.16% for Top-5 error. Similarly, for the AlexNet model, it reaches 99.48% for Top-1 error and 92.84% for Top-5 error, and for the RegNetY 400MF model, it reaches 97.08% for Top-1 error and 76.04% for Top-5 error. Among these three models, ResNet-101 appears to be the most resilient to the FGSM attack among the three models studied in this work.

**(3)** *Classification in the presence of the adversarial attack:* As mentioned in Section I, an alternative attack method on image classification models involves the use of adversarial patches [11]. In this attack, the strategy is to manipulate a small portion of the image to take on a specific form and appearance, rather than the FGSM's approach of making slight pixel changes. This manipulation aims to deceive the classification model and compel it to predict a predetermined class. In practical scenarios, this type of attack presents a more significant threat than FGSM. For example, consider an autonomous vehicle network that receives real-time images from a camera. To trick the vehicle into misidentifying an automobile as a pedestrian, another driver could print and attach a particular pattern on the rear of the vehicle.

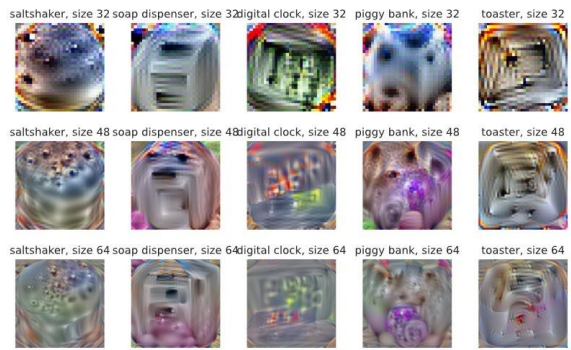

Fig. 3. Five images used as patches: (i) saltshaker, (ii) soap dispenser, (iii) digital clock, (iv) piggy bank, and (v) toaster. The sizes for the patch images are 32*32, 48*48, and 64*64.

TABLE VII. THE ACCURACY OF THE PATCH ATTACK ON THE RESNET-101 CLASSIFICATION MODEL

| Image Class | Top 1 % Accuracy | | |
|---|---|---|---|
| | Patch: 32*32 | Patch: 48*48 | Patch: 64*64 |
| saltshaker | 72.52 | 96.26 | 99.34 |
| soap dispenser | 16.47 | 78.51 | 92.65 |
| digital clock | 55.35 | 96.79 | 99.37 |
| piggy bank | 78.03 | 95.54 | 99.17 |
| toaster | 48.69 | 90.48 | 98.58 |

TABLE VIII. THE ACCURACY OF THE PATCH ATTACK ON THE RESNET-101 CLASSIFICATION MODEL

| Image Class | Top 5 % Accuracy | | |
|---|---|---|---|
| | Patch: 32*32 | Patch: 46*48 | Patch: 64*64 |
| **saltshaker** | 90.25 | 99.79 | 99.99 |
| **soap dispenser** | 61.76 | 95.77 | 99.25 |
| **digital clock** | 94.11 | 99.95 | 100.00 |
| **piggy bank** | 92.27 | 99.59 | 99.98 |
| **toaster** | 72.02 | 98.12 | 99.93 |

To simulate the adversarial patch attack on the same four images previously subjected to the FGSM attack, the process begins by randomly selecting five images to be used as the patches. These patch images include (i) a saltshaker, (ii) a soap dispenser, (iii) a digital clock, (iv) a piggy bank, and (v) a toaster. To investigate how the patch sizes affect the classification model's accuracy, three different sizes are considered for each patch image: (i) 32x32, (ii) 48x48, and (iii) 64x64, with size measurements expressed in terms of pixels along the *x* and *y* dimensions. Tables VII and VIII provide the accuracies (Top 1% and Top 5%) of the ResNet-101 model for various sizes of different patch images. Accuracy, in this context, indicates the percentage of cases in which the images are classified as the target class (i.e., the patch class) with the highest confidence.

TABLE IX. THE ACCURACY OF THE PATCH ATTACK ON THE ALEXNET CLASSIFICATION MODEL

| Image Class | Top 1 % Accuracy | | |
|---|---|---|---|
| | Patch: 32*32 | Patch: 48*48 | Patch: 64*64 |
| saltshaker | 16.85 | 54.84 | 86.40 |
| soap dispenser | 7.61 | 17.27 | 52.05 |
| digital clock | 25.70 | 69.95 | 94.99 |
| piggy bank | 1.32 | 79.58 | 96.70 |
| toaster | 48.89 | 90.48 | 98.58 |

TABLE X. THE ACCURACY OF THE PATCH ATTACK ON THE ALEXNET CLASSIFICATION MODEL

| Image Class | Top 5 % Accuracy | | |
|---|---|---|---|
| | Patch: 32*32 | Patch: 46*48 | Patch: 64*64 |
| saltshaker | 51.56 | 85.25 | 95.74 |
| soap dispenser | 36.57 | 56.38 | 86.50 |
| digital clock | 54.60 | 91.21 | 98.89 |
| piggy bank | 11.07 | 92.16 | 99.22 |
| toaster | 72.02 | 98.12 | 99.93 |

TABLE XI. THE ACCURACY OF THE PATCH ATTACK ON THE REGNETY 400MF CLASSIFICATION MODEL

| Image Class | Top 1 % Accuracy | | |
|---|---|---|---|
| | Patch: 32*32 | Patch: 48*48 | Patch: 64*64 |
| saltshaker | 23.99 | 62.95 | 93.69 |
| soap dispenser | 25.97 | 77.38 | 91.37 |
| digital clock | 34.77 | 83.49 | 93.33 |
| piggy bank | 37.00 | 81.82 | 94.24 |
| toaster | 48.89 | 90.48 | 98.58 |

TABLE XII. THE ACCURACY OF THE PATCH ATTACK ON THE REGNETY 400MF CLASSIFICATION MODEL

| Image Class | Top 5 % Accuracy | | |
|---|---|---|---|
| | Patch: 32*32 | Patch: 48*48 | Patch: 64*64 |
| saltshaker | 48.51 | 81.75 | 99.03 |
| soap dispenser | 51.13 | 90.65 | 98.24 |
| digital clock | 64.18 | 95.92 | 99.28 |
| piggy bank | 59.41 | 92.72 | 98.71 |
| toaster | 72.02 | 98.12 | 99.93 |

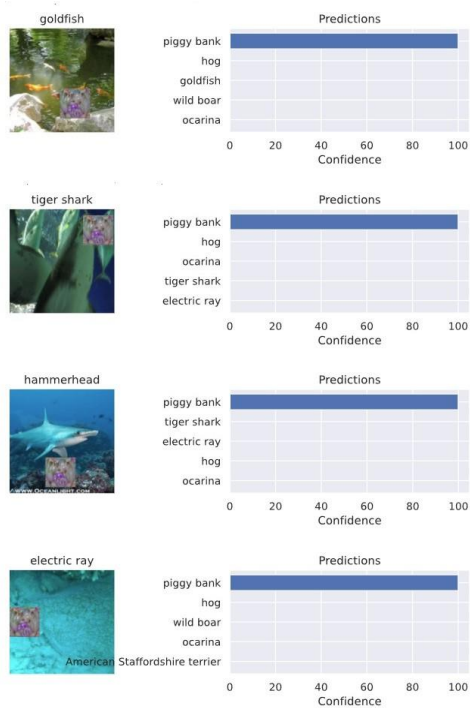

Fig. 4. The classification results of the ResNet-101 model in the presence of a patch of "piggy bank" of size 64*64.

Tables IX and X provide the accuracies (Top 1% and Top 5%) of the AlexNet model for various sizes of different patch images. The Top 1% and Top 5% accuracies for the patch attack on the RegNetY 400MF model are presented in Tables XI and XII, respectively.

Figures 4-6 illustrate the performance of the three classification models when confronted with a "piggy bank" patch sized at 64x64 since this patch image has yielded the highest accuracy for the patch attack. Due to the space constraint, the performance results of the patch attack images for other patch types and sizes are not included here.

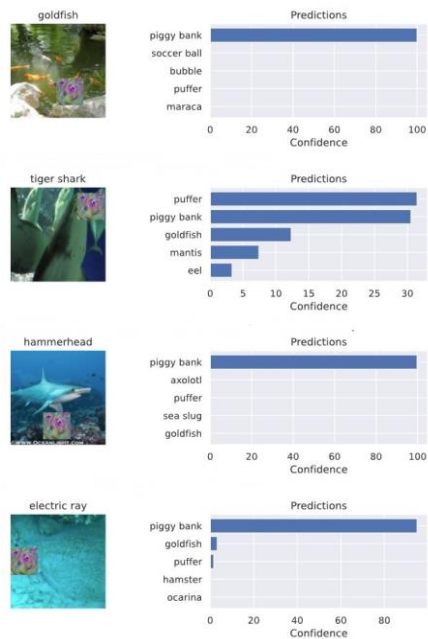

Fig. 5. The classification results of the AlexNet model in the presence of a patch of "piggy bank" of size 64*64.

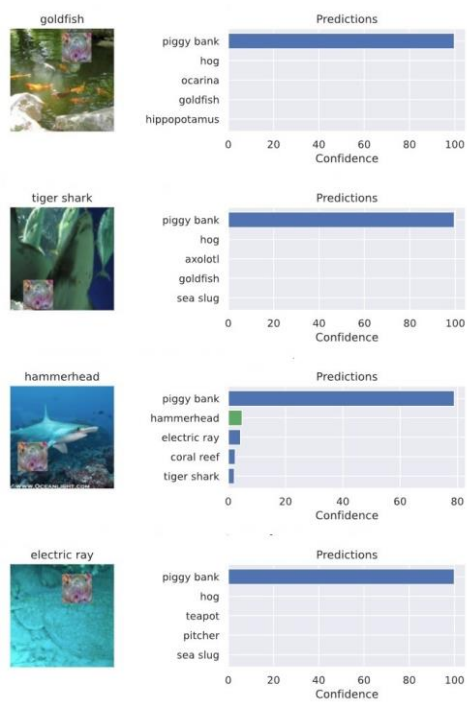

Fig. 6. The classification results of the RegNetY 400MF model in the presence of a patch of "piggy bank" of size 64*64.

It is thus observed that the patch attack has a disastrous effect on the classification performance of all three models. For most of the patch images and patch sizes, the effectiveness of the attack on three models is found to be most adverse for the patch image "toaster". Also, for the same patch, the performances of all three models have been the worst for the biggest patch of size 64*64.

## V. Defensive Distillation Method

Distillation is an approach initially created to train a Deep Neural Network (DNN) by transferring knowledge from another DNN. The idea was first proposed in [31], and the formal introduction of distillation came in [32]. The primary goal of distillation's knowledge transfer is to reduce the computational complexity of DNN models by transferring knowledge from larger, more complex architectures to smaller ones. This makes it feasible to apply deep learning on devices with limited resources, such as smartphones, which may lack powerful GPUs for computation. Papernot et al. proposed a novel variation of distillation for defensive training, where instead of transferring knowledge between distinct architectures, the authors suggested using the knowledge extracted from a DNN to enhance its resistance against adversarial inputs [33]. The authors employed the knowledge acquired through distillation to diminish the magnitude of network gradients that adversaries exploit when creating adversarial samples. It is argued that when adversarial gradients are substantial, it becomes easier to create adversarial samples because even minor alterations can lead to significant changes in the output of the DNN. To counteract such perturbations, reducing the variations around the input data is necessary, thereby decreasing the magnitude of adversarial gradients [34]. In essence, the authors used defensive distillation to refine the model learned by a DNN during training, aiding the model in better generalization to samples that differ from its training data. During testing, models trained using defensive distillation exhibit reduced sensitivity to adversarial samples, making them more suitable for deployment in security-sensitive environments.

This concept is rooted in the idea that the knowledge acquired by DNNs during their training is not limited to just the weight parameters they learn. It is also embedded within the probability distributions generated by the network. Consequently, distillation involves extracting class-related information from these probability distributions and transferring it to a different DNN architecture during training. To facilitate this transfer, distillation assigns labels to the inputs in the training dataset of the second DNN based on their classification predictions as per the first DNN. The advantage of using class probabilities instead of strict labels is apparent, as probabilities convey additional information about each class beyond simply indicating the correct class for a given sample.

In the defensive distillation process, "temperature" refers to a hyperparameter that is used to control the smoothness or softness of the probability distribution generated by the neural network's *softmax* function [33]. The softmax function is typically used in the output layer of a neural network to convert raw scores or logits into probability distributions over multiple classes. When a higher temperature value is used, such as 2 or 3, during the softmax operation, it has the effect of smoothing out the probability distribution. This means that the model becomes less confident in its predictions, and the predicted probabilities for all classes become more uniform. In other words, the model assigns relatively equal probabilities to multiple classes, making the distribution softer. On the other hand, when a lower temperature value is in place, such as 0.5 or 1, during the softmax operation, it sharpens the probability distribution. In this case, the model becomes more confident in its predictions, and one or a few classes receive significantly higher probabilities while others receive very low probabilities. The distribution becomes peakier, indicating high confidence in specific class predictions. Usually, a higher temperature setting is used when training the distilled model to mimic the teacher (i.e., the complex DNN) model. This softens the predicted probability distribution of the distilled model and makes it more resistant to adversarial attacks. The idea is that adversarial perturbations are less likely to significantly impact the model's predictions when the distribution is soft and spread out.

Hence, the distilled model can then be used as a more robust and secure alternative to the teacher model. Adversarial examples that successfully fool the teacher model may not be as effective against the distilled model because it has learned a smoother decision boundary. The intuition behind defensive distillation is that the softer, smoother decision boundaries learned by the distilled model are less susceptible to small, imperceptible perturbations in the input data, which are often characteristic of adversarial examples.

## VI. Results of Defensive Distillation Method

The MNIST dataset containing 60000 grayscale images of handwritten digits has been used for studying the effect of the FGSM attack and then the effectiveness of the defensive distillation method in defending against the attack [35]. An image classification model based on the CNN architecture is first trained on randomly chosen 50,000 samples and then tested on the remaining 10,000 samples. The CNN model consists of the following layers.

*Convolutional layer 1:* The first convolutional layer has the following details. Input: 1 channel (grayscale image), Output: 32 feature maps, Kernel size: 3*3, Stride:1, Activation function: ReLU.

*Convolutional layer 2:* The second convolutional layer has the following details. Input: 32 feature maps (from the previous layer), Output: 64 feature maps, Kernal size: 3*3, Stride: 1, Activation function: ReLU.

*Dropout layer 1:* The first dropout layer applies a 2D dropout with a probability of 0.25.

*Dropout layer 2:* The second dropout layer applies a 2D dropout probability of 0.5.

*Fully connected layer 1:* The fully connected (also known as the dense) layer has the following details. Input: the flattened output from the previous layer (9216 dimensions), Output: 128 units, Activation: ReLU.

The network is trained using the *adam* optimizer. The learning rate hyperparameter is set at a very low value of 0.0001, and the beta parameters of the *adam* optimizer are set to (0.9, 0.999). These parameters control the exponential moving averages of the gradients and the squared gradients, respectively, and are used to adaptively adjust the learning rates for each parameter. The negative log-likelihood function is used as the loss function. The "ReduceLROnPlateau" learning rate scheduler monitors the validation loss and reduces the learning rate when the monitored metric stops improving and exhibits a plateau. The "mode" parameter of the scheduler is set to "min" so that the learning rate is reduced when the monitored metric, validation loss, stops decreasing. The value of the "factor" parameter of the scheduler is set to a value of 0.1. This value specifies the learning rate will be multiplied by a factor of 0.2 when the monitored metric, validation loss experiences a plateau. The value of the "patience" parameter of the scheduler is set to 3 indicating that if the validation loss does not improve for 3 consecutive epochs, the learning rate is reduced. The model is trained over 10 epochs. Fig. 7 exhibits how the training and validation loss varied with the number of epochs during the model training.

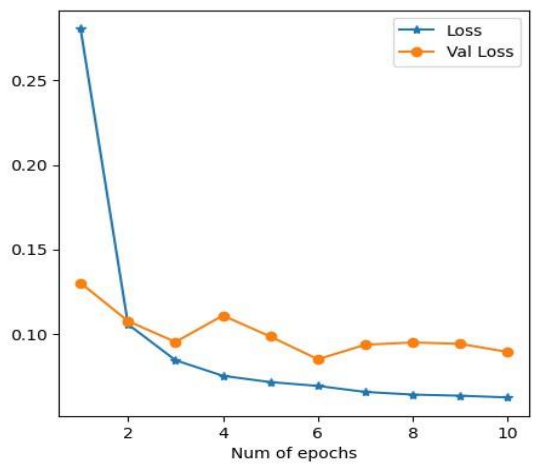

Fig. 7. The variations of the training loss and validation loss for the number of epochs for the training of the classifier model

TABLE XIII. THE EFFECT OF THE FGSM ATTACK ON THE CLASSIFICATION ACCURACY OF THE IMAGE CLASSIFIER MODEL

| Epsilon | No. of correctly classified cases | Total No of cases | Classification Accuracy |
|---|---|---|---|
| 0.000 | 9698 | 10000 | 0.9698 |
| 0.007 | 9650 | 10000 | 0.9650 |
| 0.010 | 9625 | 10000 | 0.9625 |
| 0.020 | 9585 | 10000 | 0.9585 |
| 0.030 | 9525 | 10000 | 0.9525 |
| 0.050 | 9366 | 10000 | 0.9366 |
| 0.100 | 8556 | 10000 | 0.8556 |
| 0.200 | 5405 | 10000 | 0.5405 |
| 0.300 | 2727 | 10000 | 0.2767 |

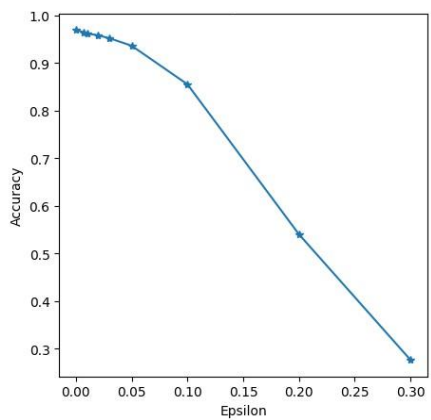

Fig. 8. The effect of the FGSM attack on the classification accuracy of an image classifier for different noise levels.

The FGSM attack is launched on the trained network by creating adversarial samples for different epsilon values. The epsilon values chosen in this study are 0.000, 0.007, 0.010, 0.020, 0.030, 0.050, 0.100, 0.200, and 0.300. Further higher values of epsilon are not considered as FGSM is effective even for smaller values of epsilon. The accuracy values of the classifier for different values of epsilon under the FGSM attack are presented in Table XIII. It is observed that in the absence of the FGSM attack, the classifier exhibited 96.88% accuracy which consistently reduced to 27.27% as the value of the epsilon parameter (i.e., the noise percentage) of the FGSM increased to 30%. Fig. 8 depicts the effect of the noise level of the FGSM attack on the classification accuracy of the classifier.

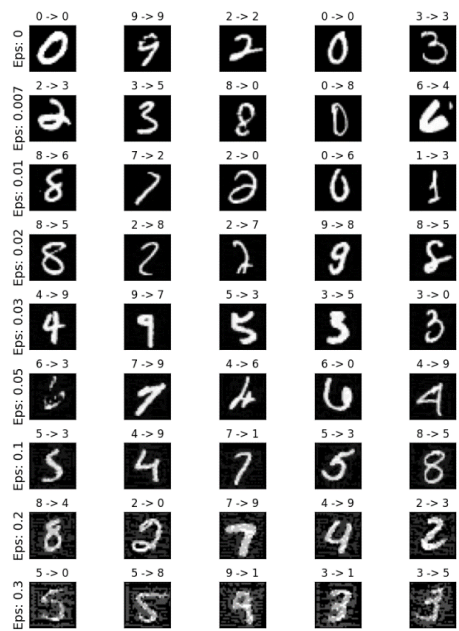

Fig. 9. The classification results for some specific images for different noise levels of the FGSM attack.

The defensive distillation method is now applied with the initial temperature set to 100, the target labels are converted to soft labels, and the distilled network is trained on the updated training data. The training and validation loss plot for the distilled network is exhibited in Fig. 10.

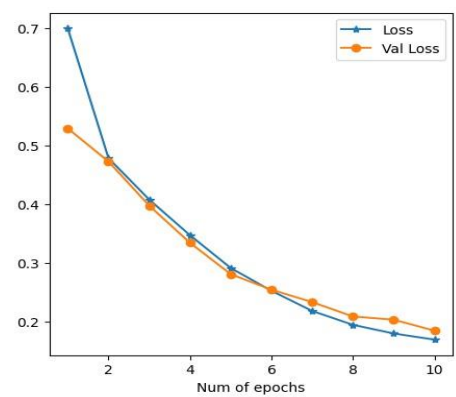

Fig. 10. The variations of the training loss and validation loss for the number of epochs for the training of the distilled network

The distilled network is now used for classifying the test cases under the presence of the FGSM attack for different values of the epsilon parameter to check the robustness of the defensive distillation method. The same values of epsilon are used which were used to study the adverse effect of the FGSM attack on the classifier in the absence of any defense mechanism. Table XIV presents the results. Fig. 11 depicts the effect of the noise level of the FGSM attack on the classification accuracy of the distilled classifier model. The distilled network is robust in defending against the adversarial samples created by the FGSM attack. The classification accuracy of the distilled network even at 30% noise level is found to be as high as 92.91%.

TABLE XIV. THE EFFECT OF THE FGSM ATTACK ON THE CLASSIFICATION ACCURACY OF THE DISTILLED CLASSIFIER MODEL

| Epsilon | No. of correctly classified cases | Total No of cases | Classification Accuracy |
|---|---|---|---|
| 0.000 | 9291 | 10000 | 0.9291 |
| 0.007 | 9275 | 10000 | 0.9275 |
| 0.010 | 9265 | 10000 | 0.9265 |
| 0.020 | 9275 | 10000 | 0.9275 |
| 0.030 | 9246 | 10000 | 0.9246 |
| 0.050 | 9216 | 10000 | 0.9216 |
| 0.100 | 9211 | 10000 | 0.9211 |
| 0.200 | 9131 | 10000 | 0.9131 |
| 0.300 | 9291 | 10000 | 0.9111 |

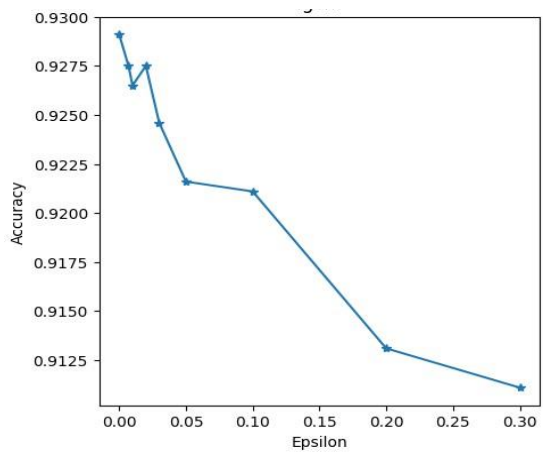

Fig. 11. The effect of the FGSM attack on the classification accuracy of the distilled network for different noise levels.

## VII. CONCLUSION

The work presented in this paper delved into the topic of adversarial attacks targeting CNN-based image classification models. Specifically, it presented a detailed study of two specific attack methods: the FGSM attack and the adversarial patch attack. The FGSM attack operates by altering the pixels of an image in the direction of their steepest gradients, maximizing the loss function value. The resulting adversarial image appears indistinguishable from the original image to the human eye, yet well-trained classification models tend to misclassify it compared to its ground truth class. In the case of the adversarial patch attack, a patch from a different class is strategically inserted into the original image, deceiving the trained models into incorrectly classifying the original image as belonging to the class of the patch. The study revealed that as the level of perturbation introduced by the FGSM attack increases in the original image, classification errors also increase until reaching a saturation point. Beyond this threshold, further perturbation typically results in a decline in classification accuracy. As for the adversarial patch attack, its effectiveness rises proportionally with the size of the patch. Finally, a mechanism for defending against the FGSM attack based on distillation of deep neural networks was presented and the effectiveness and robustness of the defense mechanism were also demonstrated. However, it is essential to note that defensive distillation is not a foolproof defense against adversarial attacks. There have been several criticisms and limitations associated with this technique. Some of these drawbacks are as follows. First, defensive distillation was shown to be effective against early adversarial attacks, such as the FGSM attack. Still, it is not as robust against more advanced attack methods, such as the Carlini-Wagner attack, which specifically aims to bypass defensive distillation [36]. Second, the process of training a distilled model involves training both a teacher and a student model, making it computationally expensive. This cost can be prohibitive for many practical applications, especially when considering the need for retraining in the face of evolving attack methods. Third, distillation can make the model more complex and less interpretable, making it challenging to understand and debug the model's behavior. Fourth, defensive distillation can make models more robust to specific types of perturbations but might not provide a holistic solution for model robustness. It may not account for all possible adversarial scenarios. Finally, over time, as adversarial attack techniques evolve, defensive distillation may become less effective at defending against these attacks. In future work, the vulnerabilities of the distillation mechanism as a defense against the FGSM attack will be explored considering the well-known Carlini-Wagnor attack [36-43].

REFERENCES

[1] C. Szegedy, W. Zaremba, I. Sutskever, J. Bruna, D. Erhan, I. J. Goodfellow, and R. Fergus, "Intriguing properties of neural networks," *Proceedings of International Conference on Learning Representations (ICLR'14)*, Poster Track, April 14-16, 2014, Banff, Canada. DOI: 10.48550/arXiv.1312.6199.

[2] I. J. Goodfellow, J. Shlens, and C. Szegedy, "Explaining and harnessing adversarial examples," *Proceedings of International Conference on Learning and Representations (ICLR'15)*, May 7-9, 2015, San Diego, CA, USA. DOI: 10.48550/arXiv.1412.6572.

[3] N. Srivastava, G. Hinton, A. Krizhevsky, I. Sutskever, and R. Salakhutdinov, "Dropout: A simple way to prevent neural networks from overfitting," *The Journal of Machine Learning Research*, Vol 15, No 1, pp. 1929-1958, 2014.

[4] NVIDIA, Solutions for Self-Driving Cars, 2023. Available online at: https://www.nvidia.com/en-us/self-driving-cars (Accessed on May 9, 2023)

[5] D. Ciresan, U. Meier, J. Masci, and J. Schmidhuber, "Multi-column deep neural network for traffic sign classification," *Neural Networks*, Vol 32, No August 2012, pp. 333-338. DOI: 10.1016/j.neunet.2012.02.023.

[6] L. Huang, A. D. Joseph, B. Nelson, B. I. P. Rubinstein, and J. D. Tygar, "Adversarial machine learning," *Proceedings of the 4th ACM Workshop on Security and Artificial Intelligence*, Chicago, IL, USA, October 21, 2011, pp. 43-58. ACM Press, New York, NY, USA. DOI: 10.1145/2046684.2046692.

[7] B. Biggio, G. Fumera, and F. Roli, "Pattern recognition systems under attack: Design issues and research challenges," *International Journal of Pattern Recognition and Artificial Intelligence*, Vol 28, No 7, Article ID: 1460002, 2014. DOI: 10.1142/S0218001414600027.

[8] B. Biggio, I. Corona, D. Maiorca, B. Nelson, N. Srndic, P. Laskov, G. Giacinto, and F. Roli, "Evasion attacks against machine learning at test time," in: H. Blockeel et al. (eds) *Machine Learning and Knowledge Discovery in Databases. ECML PKDD 2013*. Lecture Notes in Computer Science, Vol 8190, pp 387-402 Springer, Berlin, Heidelberg, Germany. DOI: 10.1007/978-3-642-40994-3_25.

[9] A. Anjos and S. Marcel, "Counter-measures to photo attacks in face recognition: A public database and a baseline," *Proceedings of the 2011 International Joint Conference on Biometrics (IJCB)*, October 11-13, 2011, Washington DC, USA, pp1-7. IEEE, 2011. DOI: 10.1109/IJCB.2011.6117503.

[10] P. Fogla and W. Lee, "Evading network anomaly detection systems: Formal reasoning and practical techniques," *Proceedings of the 13th ACM Conference on Computer and Communications Security*, October

30 – November 3, 2006, Alexandria, VA, USA, pp 59-68. ACM, New York, USA. DOI: 10.1145/1180405.1180414.

[11] T. B. Brown, D. Mane, A. Roy, M. Abadi, and J. Gilmer, "Adversarial Patch," *Proceedings of the 31st Conference on Neural Information Processing Systems (NIPS'17) Workshop*, December 4-9, 2017, Long Beach, CA, USA. Publisher: Curran Associates Inc, Red Hook, NY, USA. DOI: 10.48550/arXiv.1712.09665.

[12] S.-M. M.-Dezfooli, A. Fawzi, and P. Frossard, "Deepfool: A simple and accurate method to fool deep neural networks," *Proceedings of 2016 IEEE Conference on Computer Vision and Pattern Recognition (CVPR'16)*, June 27-30, 2016, Las Vegas, NV, USA, pp 2574-2582. IEEE Press, Piscataway, NJ, USA. DOI: 10.1109/CVPR.2016.282.

[13] A. Madry, A. Makelov, L. Schmidt, D. Tsipras, and A. Vladu, "Towards deep learning models resistant to adversarial attacks," *Proceedings of International Conference on Learning Representations (ICLR'18)*, Poster Track, April 30-May 3, 2018, Vancouver, BC, Canada. DOI: 10.48550/arXiv.1706.06083.

[14] N. Papernot, P. McDaniel, S. Jha, M. Fredrikson, Z. B. Celik, and A. Swami, "The limitations of deep learning in adversarial settings," *Proceedings of 2016 IEEE European Symposium on Security and Privacy (EuroS&P'16)*, March 21-24, 2016, Saarbruecken, Germany, pp. 372-387. IEEE Press, Piscataway, NJ, USA. DOI: 10.1109/EuroSP.2016.36.

[15] M. Sharif, S. Bhagavatula, L. Bauer, and M. K. Reiter, "Accessorize to a crime: Real and stealthy attacks on state-of-the-art face recognition," *Proceedings of the 2016 ACM SIGSAC Conference on Computer and Communications Security*, October 24-28, 2016, Vienna, Austria, pp 1528–1540. ACM Press, New York, NY, USA, 2016. DOI: 10.1145/2976749.2978392.

[16] J. Deng, W. Dong, R. Socher, L-J. Li, K. Li, and L.F-Fei, "ImageNet: A large-scale hierarchical image database," *Proceedings of 2009 IEEE Conference on Computer Vision and Pattern Recognition (CVPR'09)*, June 20-25, Miami, FL, USA, pp 248-255. IEEE Press, Piscataway, NJ, USA. DOI: 10.1109/CVPR.2009.5206848.

[17] K. Eykholt, I. Evtimov, E. Fernandes, B. Li, A. Rahmati, C. Xiao, A. Prakash, T. Kohno, and D. Song, "Robust physical-world attacks on deep learning visual classification," *Proceedings of 2018 IEEE/CVF Conference on Computer Vision and Pattern Recognition (CVPR'18)*, June 18-23, 2018, Salt Lake City, UT, USA, pp 1625-1634. IEEE Press, Piscataway, NJ, USA. DOI: 10.1109/CVPR.2018.00175.

[18] F. Tramer, A. Kurakin, N. Papernot, I. Goodfellow, D. Boneh, P. McDaniel, "Ensemble adversarial training: Attacks and defenses," *Proceedings of International Conference on Learning Representations (ICLR'18)*, Poster Track, April 30-May 3, 2018, Vancouver, BC, Canada. DOI: 10.48550/arXiv.1705.07204

[19] S. Gu and L. Rigazio, "Towards deep neural network architectures robust to adversarial examples," *Proceedings of International Conference on Learning and Representations (ICLR'15)*, Poster Track, May 7-9, 2015, San Diego, CA, USA. DOI: 10.48550/arXiv.1412.5068.

[20] K. Chalupka, P. Perona, and F. Eberhardt, "Visual causal feature learning", *Proceedings of the 31st Conference on Uncertainty in Artificial Intelligence*, Amsterdam, Netherlands, pp. 181-190, July 12-16, 2015, AUAI Press, Arlington, VA, USA. DOI: 10.48550/arXiv.1412.2309.

[21] Y. Dong, Q-A, Fu, X. Yang, T. Pang, H. Su, Z. Xiao, and J. Zhu, "Benchmarking adversarial robustness on image classification", *Proceedings of 2020 IEEE/CVF Conference on Computer Vision and Pattern Recognition (CVPR'20)*, June 13-19, 2020, Seattle, WA, USA, pp 318-328. DOI: 10.1109/CVPR42600.2020.00040.

[22] Y. Chen, M. Zhang, J. Li, and X. Kuang, "Adversarial attacks and defenses in image classification: A practical perspective", *Proceedings of the 7th International Conference on Image, Vision and Computing (ICIVC'22)*, Xian, China, July 26-28, 2022, pp 424-430. DOI: 10.1109/ICIVC55077.2022.9886997.

[23] C. Pestana, N. Akhtar, W. Liu, D. Glance, and A. Mian, "Adversarial attacks and defense on deep learning classification models using YCbCr color images", *Proceedings of 2021 International Joint Conference on Neural Networks (IJCNN'21)*, July 18-22, 2021, Shenzhen, China, pp 1-9. DOI: 10.1109/IJCNN52387.2021.9533495.

[24] C. Li, C. Fan, J. Zhang, C. Li, and Y. Teng, "A block gray adversarial attack method for image classification neural network," *Proceedings of 2022 IEEE 24th International Conference on High Performance Computing & Communications (HPCC'22)*, December 18-20, 2022, Hainan, China, pp 1682-1689. DOI: 10.1109/HPCC-DSS-SmartCity-DependSys57074.2022.00255.

[25] H. Yuan, S. Li, W. Sun, Z. Li, X. Steven, "An efficient attention-based image adversarial attack algorithm with differential evolution on realistic high-resolution image," *Proceedings of 2021 IEEE/ACIS 20th International Fall Conference on Computer and Information Science (ICIS Fall'21)*, October 1-15, 2021, Xian, China, pp. 115-120. DOI: 10.1109/ICISFall51598.2021.9627468.

[26] Y. Xu, B. Du, and L. Zhang, "Self-attention context network: Addressing the threat of adversarial attacks for hyperspectral image classification," *IEEE Transactions on Image Processing*, Vol 30, pp 8671-8685. DOI: 10.1109/TIP.2021.3118977.NSE Website: http://www1.nseindia.com

[27] J. Sen and S. Dasgupta, "Adversarial attacks on Image classification models: FGSM and patch attacks and their impact," in: J. Sen and J. Mayer (eds) *Information Security and Privacy in the Digital World: Some Selected Topics*, IntechOpen, London, UK. ISBN: 978-1-83768-196-9. DOI: 10.5772/intechopen.112442.

[28] K. He, X. Zhang, S. Ren, and J. Sun, "Deep residual learning for image recognition," *Proceedings of the 2016 IEEE Conference on Computer Vision and Pattern Recognition (CVPR)*, Las Vegas, NV, USA, 2016, pp. 770-778. DOI: 10.1109/CVPR.2016.90.

[29] A. Krizhevsky, I. Sutskever, and G. Hinton, "ImageNet classification with deep convolutional networks," *Communications of the ACM*, Vol 60, No 6, pp. 84-90, June, 2017. DOI: 10.1145/3065386.

[30] I. Radosavovic, R. P. Kosaraju, R. Girshick, K. He and P. Dollár, "Designing network design spaces," *Proceedings of 2020 IEEE/CVF Conference on Computer Vision and Pattern Recognition (CVPR)*, Seattle, WA, USA, 2020, pp. 10425-10433. DOI: 10.1109/CVPR42600.2020.01044.

[31] J. Ba and R. Caruana, "Do deep nets really need to be deep?" *Advances in Neural Information Systems*, pp. 2654-2662, 2014.

[32] G. Hinton, O. Vinyals, and J. Dean, "Distilling the knowledge in a neural network," *Deep Learning and Representation Learning Workshop at NIPS 2014*, arXiv preprint arXiv:1503.02531, 2014.

[33] N. Papernot, P. McDaniel, X. Wu, S. Jha, and A. Swami, "Distillation as a defense to adversarial perturbations against deep neural networks," *2016 IEEE Symposium on Security and Privacy (SP)*, San Jose, CA, USA, 2016, pp. 582-597. DOI: 10.1109/SP.2016.41.

[34] N. Papernot, P. McDaniel, S. Jha, S., M. Fredrikson, Z. B. Celik, and A. Swami, "The limitations of deep learning in adversarial settings," *Proceedings of the 1st IEEE European Symposium on Security and Privacy*, pp. 372-387, 2016.

[35] Y. LeCun and C. Cortes, "The MNIST database of handwritten digits," 1998.

[36] N. Carlini and D. Wagner, "Towards Evaluating the Robustness of Neural Networks," *2017 IEEE Symposium on Security and Privacy (SP)*, San Jose, CA, USA, 2017, pp. 39-57. DOI: 10.1109/SP.2017.49.

[37] N. Carlini and D. Wagner, "Adversarial examples are not easily detected: Bypassing ten detection methods," *Proceedings of the 10th ACM Workshop on Artificial Intelligence and Security (AISec'17)*, pp 3-14, November 2017. DOI: 10.1145/3128572.3140444.

[38] N. Carlini and D. Wagner, "Defensive distillation is not robust to adversarial examples," *arXiv:1607.04311*, July 2016. DOI: 10.48550/arXiv.1607.04311.

[39] W. Xu, D. Evans, and Y. Qi, "Feature squeezing mitigates and detects Carlini/Wagner adversarial examples," *arXiv: 1705.10686*, May 2017. DOI: 10.48550/arXiv.1705.10686.

[40] N. Carlini and D. Wagner, "Audio adversarial examples: Targeted attacks on speech-to-text," *arXiv:1801.01944*, March 2018. DOI: 10.48550/arXiv.1801.01944.

[41] A. Athalye, N. Carlini, and D. Wagner, "Obfuscated gradients give a false sense of security: Circumventing defenses to adversarial examples'" *arXiv:1802.00420*, July 2018. DOI: 10.48550/arXiv.1802.00420.

[42] A. Athalye and N. Carlini, "On the robustness of the CVPR 2018 white-box adversarial example defenses," *arXiv: 1804.03286*, April 2018. DOI: 10.48550/arXiv.1804.03286.

[43] A. McCarthy, E. Ghadafi, P. Andriotis, and P. Legg, "Defending against adversarial machine learning attacks using hierarchical learning : A case study on network traffic attack classification," *Journal of Information Security and Applications*, Vol 72, February 2023. DOI: 10.1016/j.jisa.2022.103398